\documentclass[journal,twoside]{IEEEtran}

\usepackage{amssymb}
\usepackage{latexsym}
\usepackage{amsmath,amssymb,amsfonts}
\usepackage{algorithmic}
\usepackage{graphicx}
\usepackage{textcomp}
\usepackage{booktabs}
\usepackage{colortbl}
\usepackage{tabularx}
\usepackage{multirow}
\usepackage{xcolor}
\usepackage{hyperref}
\usepackage{threeparttable}

\title{EchoFM: Foundation Model for Generalizable Echocardiogram Analysis}

\author{Sekeun Kim, Pengfei Jin, Sifan Song, Cheng Chen, Yiwei Li, Hui Ren, \\ Xiang Li, Tianming Liu, Quanzheng Li 
\thanks{Sekeun Kim, Pengfei Jin, Sifan Song, Cheng Chen, Hui Ren, Xiang Li, and Quanzheng Li are with the Center of Advanced Medical Computing and Analysis, Massachusetts General Hospital and Harvard Medical School, Boston, MA 02114, USA}
\thanks{Yiwei Li and Tianming Liu are with the School of Computing, The University of  Georgia, Athens, GA 30602, USA}}

\begin{document}
\fontsize{9.5}{11.5}\selectfont

\maketitle

\begin{abstract}
Echocardiography is the most widely used cardiac imaging modality, providing ultrasound video data to evaluate cardiac structure and function. However, the lack of labeled data poses a significant challenge in developing supervised deep learning models. Recently, foundation models based on self-supervised learning have demonstrated remarkable generalization performance across various domains. In this paper, we introduce EchoFM, a general-purpose vision foundation model for echocardiography trained on a large-scale dataset of over 20 million echocardiographic images from 6,500 patients. We propose a novel self-supervised learning framework designed to effectively capture both spatial and temporal features in periodic videos through a spatio-temporal consistent masking strategy and periodic-driven contrastive learning. The pretrained EchoFM can then be easily adapted and fine-tuned for a variety of downstream tasks, serving as a robust backbone model. We validate EchoFM through experiments across key downstream tasks in the clinical echocardiography workflow, leveraging public and multi-center internal datasets. EchoFM consistently outperforms SOTA methods, demonstrating superior generalization capabilities and flexibility. The code and checkpoints are available at: \url{https://github.com/SekeunKim/EchoFM.git}
\end{abstract}
\begin{IEEEkeywords}
Foundation model, Ultrasound, Cardiac \end{IEEEkeywords}

\section{Introduction}
\label{sec:introduction}
\IEEEPARstart{E}{chocardiography} is the most commonly used cardiac imaging modality to non-invasively evaluate cardiac structure and function. The increasing usage of echocardiography in diverse applications presents significant challenges for clinicians, who are frequently required to interpret a large volume of images within limited time. Recently, the integration of deep learning has significantly advanced the analysis of echocardiograms, improving both the speed and accuracy of image interpretation. In particular, deep learning-based techniques for automatic recognition and segmentation have streamlined the measurement and evaluation of cardiac structures and functions. Additionally, deep learning models provide decision-making support for diagnosing cardiac diseases, enhancing diagnostic accuracy and potentially improving the overall quality of patient care \cite{ghorbani2020deep, alsharqi2018artificial}. However, the performance of deep learning models is highly dependent on the quantity and quality of labeled data. Collecting such labeled data requires significant labor, time, and financial resources from trained medical professionals, making it highly limited. For instance, in echocardiography, human annotations for heart chamber segmentation are typically provided only for two key frames in the cardiac cycle: end-diastole (ED) and end-systole (ES), which represent the phases with the most significant cardiac changes. The limited availability of labeled data poses significant challenges for supervised deep learning and limits its broader application to various downstream tasks.

Foundation models (FMs) have recently emerged as a groundbreaking advancement in artificial intelligence. Pretrained on extensive datasets, these general-purpose models capture diverse features, enabling exceptional performance across various downstream tasks \cite{he2022masked,he2020momentum}. This approach is practical because collecting large volumes of images is easier than obtaining detailed annotations, allowing models to perform well with less labeled data. For instance, foundation models like CLIP \cite{radford2021learning}, DINO \cite{caron2021emerging}, and SAM \cite{kirillov2023segment} have demonstrated impressive zero-shot and few-shot capabilities in natural image tasks. Despite these successes in natural image tasks, the application of foundation models to medical imaging remains challenging due to substantial differences in texture, structure, and segmentation targets between natural and medical images.

The versatility and robust capabilities of foundation models have driven active development and applications in various medical fields, including radiology \cite{yang2024advancing}, pathology \cite{chen2024towards}, and genomics \cite{nguyen2024sequence}. The demand for FMs specifically designed for medical imaging is rapidly increasing, leading to developments such as CheXAgent \cite{chen2401chexagent} and EVA \cite{yao2024eva} for X-ray images, MIS-FM \cite{wang2023mis} for CT images, USFM \cite{jiao2024usfm} for ultrasound images, and LVM-Med \cite{mh2024lvm} for multimodal applications. However, echocardiography presents unique challenges that require specialized models, as shown in Fig. \ref{fig1} (a). First, echocardiography captures the cyclic process of ventricular contraction and relaxation, where the heart moves in a three-dimensional helical pattern \cite{lee2018three}. This complex motion causes anatomical structures to appear intermittently or become obscured across frames, making it difficult to fully capture with two-dimensional imaging. Second, echocardiography employs multiple imaging modes, such as B-mode and Color Doppler imaging, and a variety of scan views to assess different aspects of cardiac function. Incorporating these diverse modes and views into pretraining datasets is critical to ensuring foundation models are adaptable to a wide range of downstream tasks. Third, echocardiograms are often affected by quality issues, including low signal-to-noise ratios and speckle noise \cite{mitchell2019guidelines}. These unique characteristics highlight the need for a specialized framework that can effectively capture the temporal dependencies and periodic dynamics inherent in cardiac motion. This framework is crucial to develop a robust and versatile echocardiography foundation model capable of addressing these challenges and supporting diverse clinical applications.

In this paper, we introduce EchoFM, a general-purpose vision foundation model for echocardiography, trained on a large-scale dataset comprising two public datasets and an internal dataset. The internal dataset contains 286,080 videos, which represent over 20 million echocardiographic images from 6,500 patients. This dataset includes transthoracic echocardiograms (TTE) from diagnostic exams, consisting of B-mode images and Color Doppler Imaging. Additionally, the dataset covers a wide range of demographic and clinical conditions, such as hypertension, heart failure, and valvular heart diseases, contributing to diverse representations that enhance the generalizability of the model. EchoFM captures periodic patterns of cardiac cycles to effectively model temporal dependencies in echocardiographic videos. To achieve this, we propose a periodic contrastive loss integrated within the Masked Autoencoder (MAE) \cite{he2022masked} framework, using a temporal self-similarity matrix to identify positive and negative pairs. This approach ensures consistent representations for the same cardiac phase across cycles and distinct representations for different phases. To further enhance temporal consistency, we introduce a spatio-temporal consistent masking strategy. This ensures that the temporal embeddings from visible tokens remain comparable, facilitating effective optimization of the periodic contrastive loss. We demonstrate the efficiency of EchoFM by evaluating its performance in various stages of the clinical echocardiography workflow, as shown in Fig. \ref{fig1} (b). To assess its generalization capabilities, we conducted evaluations using both public datasets and internal multicenter datasets. The key contributions of our work are summarized as follows:

\begin{figure*}
	\centering
	\includegraphics[width=\textwidth]{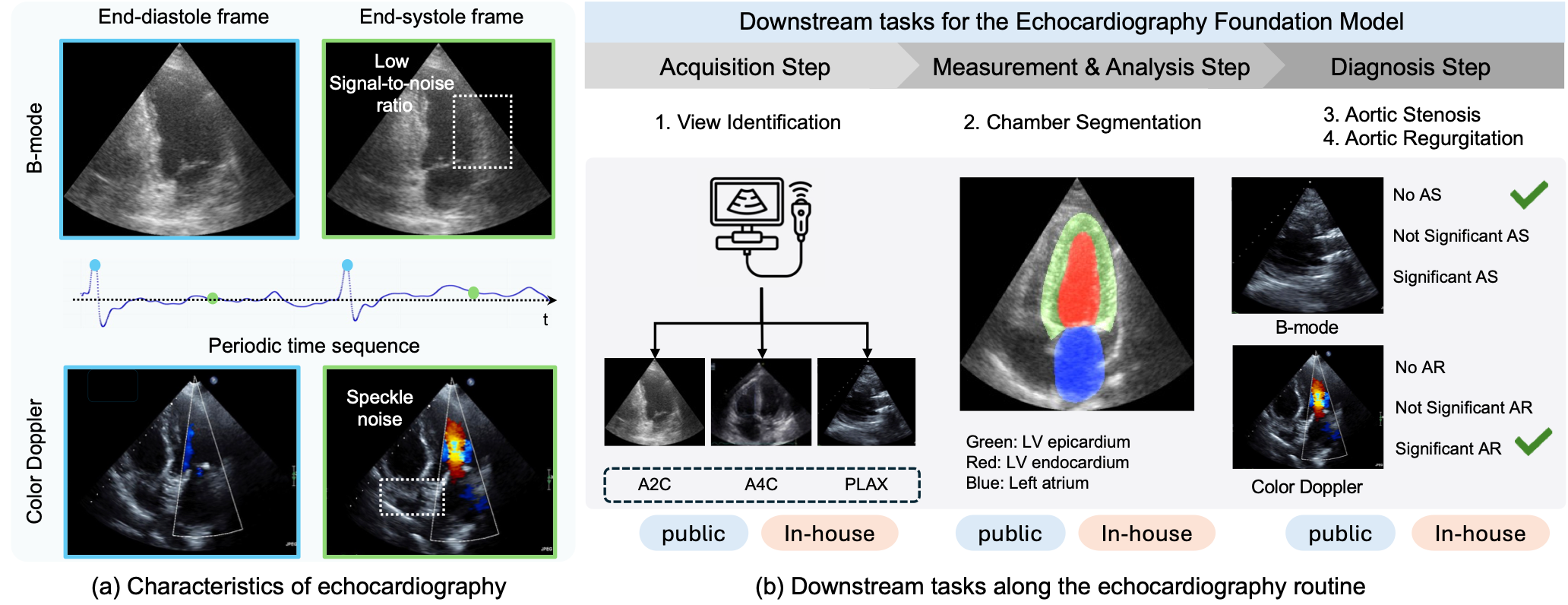}
	\caption{\textmd{(a) Key characteristics of B-mode echocardiography include its low signal-to-noise ratio and periodic temporal sequences. Unlike other imaging modalities, echocardiography features diverse scanning views and multiple imaging modes, enabling comprehensive cardiac assessment. (b) Typical downstream tasks in routine echocardiography include view identification and chamber segmentation, with a focus on disease diagnosis, such as assessing the severity of aortic stenosis and aortic regurgitation.}}
	\label{fig1}
\end{figure*}

In this paper, we introduce EchoFM, a general-purpose vision foundation model for echocardiography, trained on a large-scale dataset comprising two public datasets and an internal dataset. The internal dataset contains 286,080 videos, which represent over 20 million echocardiographic images from 6,500 patients. This dataset includes transthoracic echocardiograms (TTE) from diagnostic exams, consisting of B-mode images and Color Doppler Imaging. Additionally, the dataset covers a wide range of demographic and clinical conditions, such as hypertension, heart failure, and valvular heart diseases, contributing to diverse representations that enhance the generalizability of the model. EchoFM captures periodic patterns of cardiac cycles to effectively model temporal dependencies in echocardiographic videos. To achieve this, we propose a periodic contrastive loss integrated within the Masked Autoencoder (MAE) \cite{he2022masked} framework, using a temporal self-similarity matrix to identify positive and negative pairs. This approach ensures consistent representations for the same cardiac phase across cycles and distinct representations for different phases. To further enhance temporal consistency, we introduce a spatio-temporal consistent masking strategy. This ensures that the temporal embeddings from visible tokens remain comparable, facilitating effective optimization of the periodic contrastive loss. We demonstrate the efficiency of EchoFM by evaluating its performance in various stages of the clinical echocardiography workflow, as shown in Fig. \ref{fig1} (b). To assess its generalization capabilities, we conducted evaluations using both public datasets and internal multicenter datasets. The key contributions of our work are summarized as follows:

\begin{itemize}
\item We introduce EchoFM, a foundation model for echocardiography, trained on a large-scale dataset comprising two public datasets and an internal dataset. The internal dataset includes 286,080 videos with over 20 million images from 6,500 patients, featuring TTE with B-mode and Color Doppler imaging in diverse clinical conditions.

\item We propose a spatio-temporal consistent masking strategy for contrastive learning in videos with periodic patterns. This strategy leverages the inherent periodicity of echocardiographic sequences to enhance representation robustness by utilizing information from temporally adjacent frames when certain regions are occluded or less visible.

\item We validate EchoFM through extensive experiments with various downstream tasks across the clinical routine of echocardiograph, using both public and multi-center internal datasets. EchoFM consistently outperforms state-of-the-art (SOTA) methods, demonstrating its superior generalization capabilities and flexibility for a variety of downstream tasks.

\end{itemize}
 

The remainder of this paper is organized as follows. In Section \ref{sec:RW}, we discuss related work on self-supervised learning and foundation models. In Section \ref{sec:Method}, we present EchoFM, a foundation model specifically designed for echocardiography. Subsequently, we present and analyze the results of our extensive numerical experiments on public and internal datasets in Section \ref{sec:Experiments}. Finally, we discuss potential future directions in Section \ref{sec:discussion} and provide concluding remarks in Section \ref{sec:Conclusion}.

\section{Related Work}
\label{sec:RW}
\subsection{Self-Supervised Learning} 

Self-supervised learning methods have gained significant attention in computer vision, focusing on various pretext tasks for pre-training models without requiring manual annotations \cite{radford2021learning, caron2021emerging, chen2021exploring, grill2020bootstrap}. Some works focused on contrastive learning techniques to enhance overall feature representations, including global and patch-level characteristics. CLIP \cite{radford2021learning} uses paired text and image data to align visual and textual representations, enabling zero-shot capabilities in various tasks. DINO \cite{caron2021emerging} employs a teacher-student siamese network where the teacher’s predictions are distilled into the student network. SIMSIAM \cite{chen2021exploring} supported
the idea of neglecting the negative samples and relied only
on a Siamese network and stop-gradient operation to achieve
state-of-the-art performance. Other methods leveraged the masked image modeling (MIM) technique which employs a 'filling-in-the-blank' objective, similar to that used in large language models \cite{kenton2019bert}.
 In this framework, part of the input data is masked, and the network
is trained to reconstruct predefined targets such as pixel values \cite{he2022masked, assran2023self, bao2021beit}. Masked Autoencoder (MAE) \cite{he2022masked} demonstrated that representations learned through this strategy exhibit strong generalization capabilities when vision transformers \cite{dosovitskiy2020image} are used as the backbone architecture. More recently, I-JEPA \cite{assran2023self} introduced a novel approach involving masked feature prediction in the latent space, further enhancing self-supervised learning efficiency.

 For video-related self-supervised learning, there is an increased emphasis on temporal aspects. VideoMAE \cite{tong2022videomae} and ST-MAE \cite{feichtenhofer2022masked} extend the concept of random masking from 2D image patches to 3D spatiotemporal cubes, masking spacetime patches in videos, and training an autoencoder to reconstruct them at the pixel level. This approach has achieved promising results in video recognition by effectively capturing both spatial and temporal dynamics. There is interest in the use of self-supervised learning for periodic video data, where the focus shifts to capturing repetitive patterns. RepNet \cite{wandt2019repnet} focuses on learning discriminative features by detecting repetitive patterns, making it particularly effective for periodic medical data, such as cardiac cycles. SimPer \cite{yang2022simper} employs frequency-based augmentations to capture robust representations of periodic motion, further enhancing the model’s ability to process periodic video data effectively. Although methods like VideoMAE and ST-MAE capture general spatio-temporal dynamics, integrating periodic-specific information is essential for effectively modeling repetitive patterns and periodic dynamics, particularly in echocardiography.

\subsection{Vision Foundation Model}
 Building on the strong generalizability of self-supervised learning, foundation models have become a key approach to computer vision, using large-scale datasets and pretraining to learn representations that capture diverse and invariant features across various tasks. With the vast amount of image and text pairs available on the Internet, CLIP \cite{radford2021learning} uses a contrastive learning technique to align images and text, allowing it to learn multi-modal shared representations in a self-supervised manner. Although its effectiveness is evident in classification and retrieval tasks, CLIP faces challenges in handling dense prediction tasks, such as object detection and segmentation, which require spatially localized information. Among the representative vision foundation models, such as SegGPT \cite{wang2023seggpt} and SEEM \cite{zou2024segment}, the Segment Anything Model (SAM) \cite{kirillov2023segment} stands out for its general-purpose image segmentation capabilities, developed using a large dataset of 11 million natural images. Its extension, SAM-2 \cite{ravi2024sam}, further expands its functionality to general-purpose video segmentation. However, these general-purpose models often face performance degradation in medical downstream tasks due to a significant domain gap, which arises from differences in visual characteristics, such as homogeneous textures and lower resolutions typical of medical imaging, as well as the presence of domain-specific artifacts like speckle noise in ultrasound images. 
 
\subsection{Medical Image Foundation Model}
To address this challenge, foundation models have been tailored for the medical imaging domain, particularly in radiology. These models span various imaging modalities, such as CT, X-ray, and ultrasound, demonstrating their adaptability to different types of medical data \cite{wang2023mis, yang2024advancing, yao2024eva, mh2024lvm, jiao2024usfm, ghesu2022contrastive}. Foundation models tailored for ultrasound imaging, an essential modality in dynamic diagnostic scenarios, have recently gained significant attention. EchoCLIP \cite{christensen2024vision}, which uses a large static 2D image data set and a pair of reports, has significantly advanced the development of foundation models focusing on echocardiography. Based on this, EchoPrime adopts the CLIP framework and incorporates a modified encoder using mVIT \cite{fan2021multiscale} to process video embeddings. By integrating video and report data, EchoPrime enables a more comprehensive understanding of echocardiography videos, facilitating a better interpretation of dynamic information. However, like other CLIP-based models, it faces limitations in handling dense prediction tasks, such as segmentation.

The ultrasound vision foundation model, USFM \cite{jiao2024usfm}, is pretrained on a large-scale multi-organ ultrasound dataset using spatial-frequency masked image modeling (MIM). It demonstrates efficacy in various tasks including image segmentation, enhancement, and classification. SonoSAM \cite{ravishankar2023sonosam} fine-tunes the SAM model for ultrasound image segmentation using a pretrained SAM's image encoder while fine-tuning mask decoder and prompt encoder. SonoSAMTrack \cite{ravishankar2023sonosamtrack} utilizes SonoSAM for 2D segmentation based on manual point prompts, combined with an independent tracking algorithm called deAoT presented in \cite{yang2022decoupling} for video segmentation. Although these models perform well on 2D segmentation tasks, they face significant challenges in capturing the complex variability and periodic temporal dynamics critical to echocardiography videos. Addressing these limitations requires a foundation model that effectively integrates spatio-temporal representations, supported by large-scale self-supervised pretraining to enhance generalization and diagnostic accuracy.

\begin{figure*}
    \centering
    \includegraphics[width=1.0\linewidth]{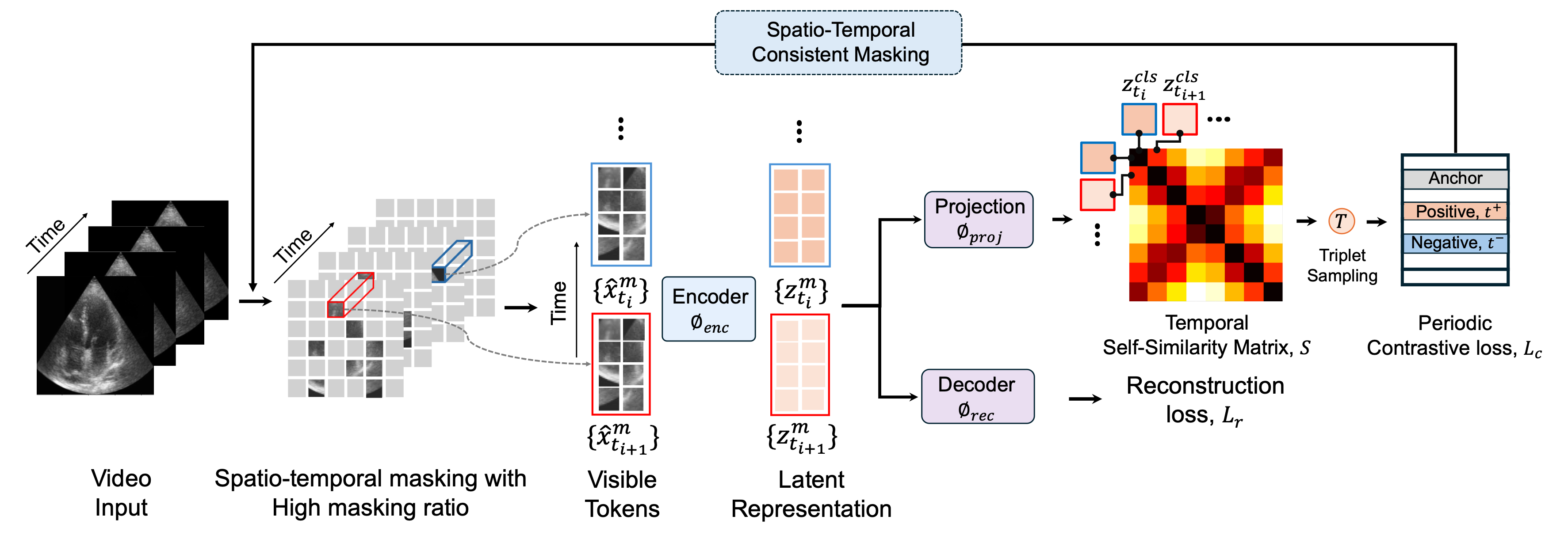}
    \caption{The overview of the proposed EchoFM. We extract spatio-temporal patches, keeping mask-ratio along temporal domain. The visible patches are processed by the ViT encoder and extracted latent representation are grouped into temporal dimension. The decoder reconstructs the missing patches in the video input. The grouped spatio-tempopral patches in temporal dimension are processed by a ViT-based projector to extract [CLS] tokens independently. We build the temporal self-similarity matrix by calculating similarity between [CLS] tokens. We sample triplet pairs, then the Spatio-temporal Consistent masking applied to triplet patches. We minimize periodic contrastive loss and reconstruction loss until network convergence. The encoder is attached to a task-specific decoder for fine-tuning and used for downstream tasks.
    }
    \label{method}
\end{figure*}

\section{Method}
\label{sec:Method}

We introduce a foundation model for echocardiography, termed EchoFM, which is based on MAE for unsupervised video training without requiring labels. EchoFM consists of an encoder that processes input video data by applying a high masking ratio to create spatio-temporal patches, a decoder for reconstructing the masked tokens, and a projection head that learns feature representations from periodic video data. The projection head serves as a feature extractor for temporal representation, effectively capturing and utilizing the periodic patterns of cardiac cycles through Spatio-temporal Consistent Masking. After pretraining on a large-scale echocardiography dataset, we freeze the EchoFM encoder and fine-tune only the adapter layers tailored to each downstream task. An overview of our proposed method is illustrated in Fig. \ref{method}.


\subsection{Encoder with Spatio-temporal Consistent Masking}
Given a video $x\in \mathbb{R}^{c\times T\times H\times W}$, where $H \times W$ represents spatial dimensions, $T$ is the number of frames, and $c$ denotes the number of input channels, which is set to 3, accommodating both B-mode and Color Doppler Imaging for consistency. We divide the input into spatio-temporal patches $x^p_{t_j,i}\in \mathbb{R}^{c\times t\times h\times w}$, where t = 16, h = 16, and w = 4. These sizes were empirically determined, as shown in Table \ref{ab:patch_size}.

To facilitate understanding, in our notation, $x$ represents videos and feature maps, $z$ denotes tokens, and $M$ stands for masks. The subscript $t_j$ indicates different times, while $i$ denotes different spatial positions. The superscript $p$ signifies the segments formed into patches, and $m$ represents the segments after masking.

The total number of spatio-temporal patches is $N =N_T \times N_H \times N_W$. Here, $N_H=H/h$, $N_W=W/w$, and $N_T=T/t$ represent the number of patches along the height, width, and temporal dimensions, respectively. $i$ spans from $1$ to $N_W\times N_T$ indexing spatial dimensions within each frame, and $t_j$ is an index ranging from $1$ to $N_T$ indicating positions in the temporal dimension. Unlike the tokenizer for 2D images, this approach not only extracts spatial features between frames but also effectively captures the motion and behavioral patterns of objects, thereby enhancing the model's understanding and predictive ability of video content.

The pathches are reshaped into a sequence of flattened tokens and projected to a size of $D$ using a 3D convolution layer. Subsequently, we add a $D$-dimensional embedding to each token using trainable position embeddings. Specifically, the embedding process is defined as: $\hat{x}^p_{t_j,i} = x^p_{t_j,i} E+E_{pos}(t_j,i),$ where $E\in R^{(c \times t\times h\times w)\times D}$ is the patch embedding projection and $E_{pos}$ represents the position embedding. In addition, we concatenate a $[CLS]$ token to embedded tokens.

To effectively learn robust features, we employ a high mask ratio of 75\% on spatial tokens and generate visible spatial tokens $\hat{x}^m_{t_j,i} = (1-M_{ij}) \odot \hat{x}^p_{t_j,i}$, where $M_{ij} \in \{0,1\}$ and $\odot$ represents the element-wise multiplication. However, to ensure that contrastive learning remains rational and effective, it is imperative to impose certain constraints on the masking procedure.

\begin{figure}[!t]
\centering
\includegraphics[width=0.495\textwidth]{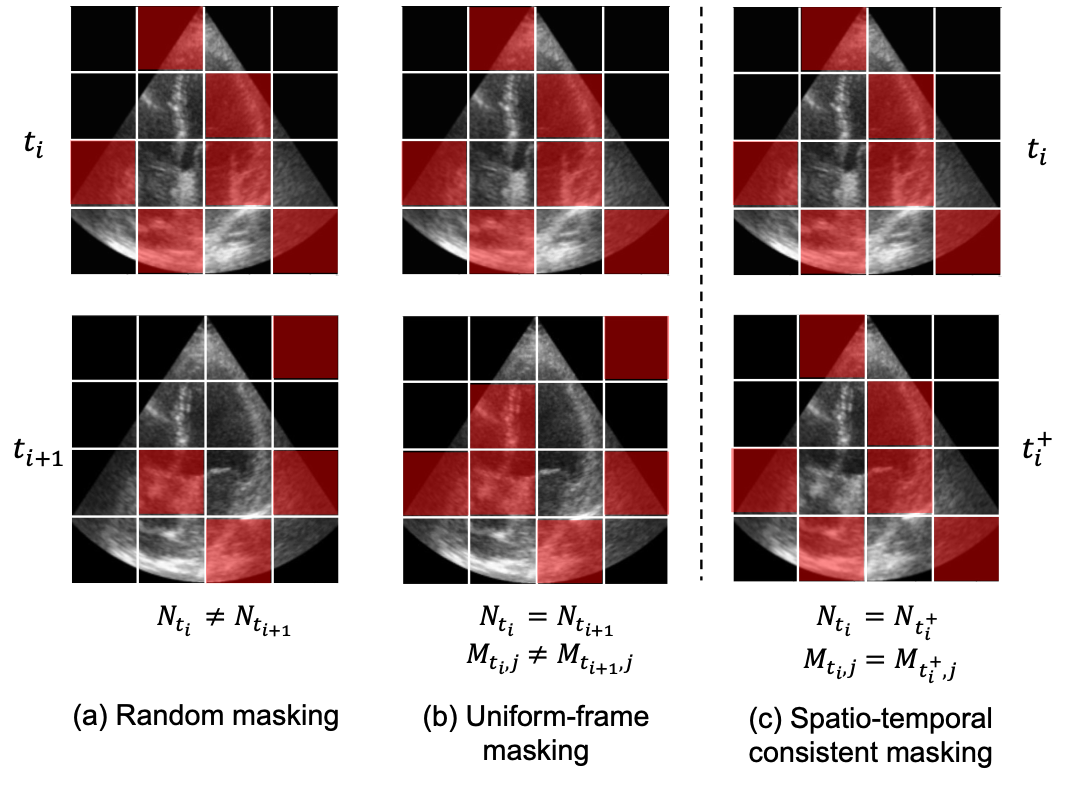}
	\caption{Different masking strategies: (a) Random masking, (b) Uniform-frame masking, which maintains the same mask ratio across the temporal dimension, and (c) Spatio-temporal consistent masking, which applies the same mask to selected samples.}
	\label{fig:visualization}
\end{figure}

Unlike existing MAE methods that employ random masking of patches in spatio-temporal data, we propose two strategic approaches: Uniform-Frame Masking and Spatio-temporal Consistent Masking, as illustrated in Fig. \ref{fig:visualization}.

The Uniform-Frame Masking applies masks to video patches while maintaining the same ratio of masked patches across each temporal dimension. By ensuring that the same proportion of patches is masked within each temporal dimension, i.e., $\sum_i M_{t_j,i}$ is constant for any $j$, we provide a balanced and effective learning process for both spatial and temporal dimensions. Additionally, this strategy facilitates contrastive learning across different time frames, thereby enhancing the model's ability to discern temporal dependencies.

The Spatio-temporal Consistent Masking is designed to enhance contrastive learning by ensuring that $M_{t_j,i} = M_{t_j^+,i} = M_{t_j^-,i}$ across the corresponding positive and negative samples of $t_j$, denoted as $t_j^+$ and $t_j^-$, respectively. The selection criteria for these samples will be discussed in subsequent sections. Maintaining consistency in the masking of these sample pairs is essential, especially given the high mask ratio. This consistency ensures that the embeddings $z^m_{t_j,i}$, $z^m_{t_j^+,i}$, and $z^m_{t_j^-,i}$ are directly comparable. This comparability is crucial for effective learning, as it compensates for the significant variation in information content present at different positions on echocardiograms.

After masking, a ViT encoder is used to obtain latent representations $\{z^m_{t_j,i}\} = \phi_{enc}(\{\hat{x}^m_{t_j,i}\})$. The visible tokens are then fed into the encoder, which consists of several attention-based ViT blocks. Each block utilizes multi-head self-attention and feed-forward neural networks to capture complex relationships and dependencies within the visible patches. The $[CLS]$ token is used to aggregate global information. The encoder extracts rich, high-dimensional representations of the video, which are crucial for the subsequent reconstruction of the masked patches, thereby enabling the model to learn the underlying structure and dynamics of the video data.


\subsection{Decoder for Spatio-temporal representation} 

To learn robust spatio-temporal representations for echocardiographic features while leveraging the cyclic nature of cardiac motion, we introduce a decoder $\phi_{rec}$ for reconstruction alongside a projection $\phi_{proj}$ for contrastive learning.

The decoder is designed to reconstruct the input video from masked observations. Specifically, encoded latent representations of visible spatial tokens, denoted as $\{z^m_{t_j,i}\}$, are concatenated with learnable masked tokens [MISS] to create combined tokens. Positional embeddings are then added to all tokens. Subsequently, this complete sequence is processed through the decoder $\phi_{rec}$, which is composed of several Transformer blocks. The final output is a sequence of reconstructed patches that are reshaped and combined to recreate the entire video. By minimizing the error between the reconstructed patches and the original patches, this framework effectively learns the underlying structure and dynamics of the video data.

As mentioned above, echocardiography captures the cyclic process of ventricular contraction and relaxation, which requires effective capture and use of temporal dependencies. To address this, we introduce a frame projection designed to learn periodic representations, denoted as $z_{t_j}^{cls}=\phi_{proj}(\{z^m_{t_j,i}\}_i)$.

More precisely, we employ a ViT with shared weights as projection head across different time frames $t_j$. In alignment with conventional ViT architectures, the learnable [CLS] tokens are concatenated with the spatial token $\{z^m_{t_j,i}\}_i$ for each $t_j$. The role of the [CLS] token is pivotal: it accumulates and integrates global information from the spatial tokens, effectively capturing the overarching temporal dynamics of each frame $t_j$. After processing through ViT, the extracted [CLS] tokens $z_{t_j}^{cls}$ should exhibit the periodic nature of $t_j$. This characteristic makes them particularly suitable for periodic contrastive learning, where understanding and distinguishing between different phases of the cardiac cycle is crucial.

\subsection{Training with periodic contrastive loss}
Our method incorporates periodic contrastive loss, designed to exploit the inherent periodicity of echocardiographic videos for robust feature representation. A prerequisite for this approach is that the training data should include at least one complete cardiac cycle, enabling the accurate identification of positive and negative frame pairs based on their respective cardiac phases.

During training, two types of loss functions are utilized. The first is the reconstruction loss, which quantifies the discrepancy between the reconstructed and the original video. The second type is the periodic contrastive loss, which measures the distance between positive and negative sample pairs. We adopt the Mean Squared Error (MSE), which is computed only on masked patches between the decoder prediction and the video input.
\begin{equation}
L_{r} = \frac{1}{N}\sum \|x_i^r - x_i^p\|^2,
\end{equation}
where $N$ represents number of mask patches in video.

For periodic contrastive loss, unlike typical contrastive learning that uses labels or data augmentation to define positive and negative sample pairs. Our assumption is that frames of the same phase throughout the cardiac cycle should exhibit similar representations. We utilize a triplet loss \cite{hoffer2015deep} to enforce periodic representations in the embedding space. The core idea is to select frames as positives if they represent the same phase as the anchor and as negatives if they represent a different phase. After the projection head, we calculate the similarity between the anchor and the remaining latent representations of each frame $z_{t_j^{cls}}$ to generate a self-similarity matrix $S$ as follows:
\begin{equation}
    \label{eq:loss1}
S_{ij} = dist(z^{cls}_{t_i}, z^{cls}_{t_j}),
\end{equation}
where $dist$ denotes $L_2$ distance between the $t_i$ and $t_j$. Our triplet loss is defined as: 
\begin{equation}
    \label{eq:loss2}
   \mathcal{L}_{c}= \sum_{i\in D}\max(\alpha, \,dist(z^{cls}_{t_i},z^{cls}_{t_i^+}) - dist(z^{cls}_{t_i},z^{cls}_{t_i^-})),
\end{equation}
where an anchor index $t_j$, we sample the triplet $\{t_j,\,t_j^{+},\,t_j^{-}\}$ by thresholding the mean of similarity matrix $thres = \frac{1}{N_h}\sum_{j=1}^{N_h}(S_{ij})$. $\alpha$ is a margin between positive and negative pairs. If similarity lower than $thres$, then we set $t_j^{+}\in \mathcal{P}_{t_j}$ and $t_j^{-}\in \mathcal{N}_{t_j}$ to be the sets of positive and negative indices, respectively. This contrastive loss encourages the representation of positive frames to be drawn closer together while pushing the negative frames away within the embedding space derived from the cardiac cycle. Therefore, it is essential that the training data include at least one complete cardiac cycle.

Given the nature of echocardiographic sequences, where adjacent frames often have similar appearances, we exclude indices adjacent to the selected anchor from the negative set. This ensures that negative samples represent truly distinct phases, enhancing the robustness of periodic representation learning.
To avoid the circular definition between the Spatio-temporal consistent Masking strategy and the similarity matrix, we use the similarity matrix obtained from the Uniform-Frame Masking to define positive and negative sample pairs. Then, we apply the Spatio-temporal Consistent Masking strategy to measure the contrastive loss.

Additionally, using multiple frames instead of a single frame spatio-temporal patch can provide benefits for learning temporal representations. Since it embeds spatio-temporal patches that consider information from multiple frames, it can learn more robust features compared to single-frame embeddings. In the ablation study, we will demonstrate that this approach improves the performance and robustness of the triplet loss for learning.

We initialize the pretrained encoder with VideoMAE \cite{tong2022videomae} spanning $100$ epochs, randomly shuffling two public datasets and an internal dataset. After initialization, the model undergoes training on the same dataset with projection heads for triplet loss. EchoFM leverages both the reconstruction loss and contrastive loss during optimization:
\begin{equation}
L_{total} = L_{r} + L_{c},
\end{equation}
\section{Experiments}
\label{sec:Experiments}
In this section, we evaluate the performance of our EchoFM model within the clinical echocardiography workflow, as illustrated in Fig. \ref{fig1}. We structured the routine echocardiography workflow into three key steps: image acquisition, measurement and analysis, and diagnosis. In the image acquisition step, we focus on view identification tasks, which involve selecting appropriate scan views from a study set. For the measurement and analysis step, we select chamber segmentation tasks. In the diagnosis step, we focus on diseases related to aortic stenosis (AS). Within this step, we divide the downstream tasks into two parts: the first task is the diagnosis of AS using grayscale B-mode imaging, and the second task is the estimation of the severity of aortic regurgitation (AR) using Color Doppler Imaging. Thus, our evaluation focused on four key downstream tasks: view identification, chamber segmentation, AS diagnosis, and AR severity estimation. We selected ViT-Large as the backbone model to maintain consistency throughout the experiments.

\begin{table}[t]
\centering
\caption{Patient Characteristics in the Multi-Center Dataset. Characteristics are listed as n (\%).}
\label{tab:clinical_characteristics}
\begin{threeparttable}
\renewcommand{\arraystretch}{1.1}
\begin{tabular}{ll}
\toprule[1.1pt]
\textbf{Characteristic}             & \textbf{Value} \\ \midrule
Patients                            & 6,494          \\
Videos                              & 286,080        \\
Age, mean $\pm$ SD $^{a}$           & 76.9 $\pm$ 10.9 \\
Female                              & 2,870 (44.2\%) \\ 
\textbf{Clinical Conditions}        &                \\
Heart Failure                       & 2,346 (36.1\%) \\
Pulmonary Hypertension              & 570 (8.7\%)    \\
Aortic Valve Stenosis               & 399 (6.1\%)    \\
Mitral Valve Stenosis               & 1,011 (15.5\%) \\
Tricuspid Valve Stenosis            & 963 (14.8\%)   \\ \bottomrule[1.1pt]
\end{tabular}
\begin{tablenotes}
\footnotesize
\item[$a$] Age is reported in years.
\end{tablenotes}
\end{threeparttable}
\end{table}

\subsection{A Large-Scale Dataset for Pretraining}

Our large-scale dataset consists of a multi-center dataset and two public datasets, ensuring diversity to support effective model pre-training. The multi-center dataset comprises 683,560 transthoracic echocardiograms (TTE) from 6,500 unique patients, collected between 2017 and 2022 from Massachusetts General Hospital (MGH) and Brigham and Women’s Hospital (BWH). The dataset includes 286,080 videos of B-mode and Color Doppler imaging in 12 standard echocardiography views \cite{madani2018fast}, collected during routine clinical examinations. The dataset consists of a wide range of demographic conditions, including hypertension, heart failure, diabetes, and valvular heart disease, as detailed in Table~\ref{tab:clinical_characteristics}. Our dataset was collected using ultrasound probes from multiple manufacturers, with GE accounting for 20\%, Philips for 78\%, and Siemens for 2\% of the data.

Preprocessing echocardiography inputs with periodicity is an important step for effective pretraining in our approach. To achieve this, we utilized lead II electrocardiogram (ECG) recorded simultaneously during echocardiography scans and detected R-R peaks \cite{Makowski2021neurokit} to identify at least one complete cardiac cycle in each video. For datasets without ECG recordings, such as the CAMUS dataset and EchoNet-Dynamic \cite{ouyang2020video}, alternative strategies were employed. Specifically, for the CAMUS dataset, which contains half-cycle videos, we applied reverse padding to construct complete cardiac cycles. Similarly, for the EchoNet-Dynamic dataset, frames were cropped from ED to ES and reverse padded to synthesize full cardiac cycles. These preprocessing methods ensure that the model effectively captures comprehensive cardiac dynamics.

\begin{table}[t]
\centering
\caption{Intraobserver Variability Analysis for the Segmentation Downstream Task.}
\renewcommand{\arraystretch}{1.1}
\begin{tabularx}{\columnwidth}{l >{\centering\arraybackslash}X >{\centering\arraybackslash}X >{\centering\arraybackslash}X}
\toprule[1.1pt]
\textbf{ROI} & \textbf{ICC} & \textbf{95\% CI} & \textbf{\textit{p}-value} \\
\hline
LV$_{endo}$ & 0.964 & 0.961–0.967 & \textless 0.001 \\
LV$_{epi}$  & 0.941 & 0.932–0.950 & \textless 0.001 \\
LA          & 0.947 & 0.940–0.954 & \textless 0.001 \\
\bottomrule[1.1pt]
\end{tabularx}
\label{repro}
\end{table}

\subsection{Diverse Downstream Datasets and Evaluation metrics}
\subsubsection{Acquisition Step - Task 1: Scan View Identification}
The Tufts Medical Echocardiogram Dataset, version 2 (TMED-2) contains 599 studies from 577 patients. Each study includes multiple 2D TTE scans, comprising a total of 5,261 2D images labeled with view and diagnosis of AS. The 2D images of each study are labeled with one of the view annotations in A2C, A4C, PLAX, PSAX, and others. Each 2D image, originally in grayscale with a resolution of 112 × 112, was converted into a three-channel pseudo-color representation by duplicating the single channel. This preprocessing step ensures compatibility with models designed for three-channel inputs. Additionally, all images underwent anonymization to protect patient privacy. We use the same data split as \cite{wessler2023automated}, which contains 17760, 3602, and 3602 frames for training (60\%), validation (20\%), and test sets (20\%), respectively. We also curated an internal view classification dataset with the same labels as the TMED-2 dataset, consisting of 250 video clips of B-mode imaging. These intrenal datasets are used exclusively for evaluation purposes.

\subsubsection{Measurement and Analysis Step - Task 2: Chamber Segmentation}
We performed chamber segmentation on two echocardiography datasets: the public CAMUS dataset \cite{leclerc2019deep} and an internal dataset. We fine-tuned the model on the public dataset and evaluated it on two datasets. The CAMUS dataset contains 2D echocardiography, comprising both apical two-chamber (2CH) and four-chamber (4CH) views of 500 patients. It provides sparse annotations along the cardiac cycle only in the ED and ES. The ground truth of three structures the left ventricle (LV$_{endo}$), the epicardium (LV$_{epi}$), and the left atrium (LA) for 2CH and 4CH are provided. Half of patients have an ejection fraction (EF) lower than 45\%, 19\% of the images are poor quality. We divided it into training, validation, and test sets in a ratio of 7:1:2. Additionally, we used a curated internal dataset to evaluate the model’s generalization performance. We collected a multi-center dataset comprising B-mode echocardiography images from 100 patients, with each patient having apical A2CH and A4CH views. The multi-center segmentation dataset was not used for pretraining EchoFM. Two experienced clinicians labeled and reviewed the delineation process. The annotations include the boundaries of the LV$_{endo}$, LV$_{epi}$, and LA during the ED and ES phases. To ensure the reliability of these annotations, an experienced clinician evaluated a randomly selected sample of 50 patients twice, with a 2-month interval between the assessments. Intra-class correlation coefficients (ICC) were calculated to evaluate the absolute agreement between the two measurements, as summarized in Table \ref{repro}. Given intermittent noise and image obscuration in the images, clinicians examined adjacent frames in the video sequences to accurately define boundaries, according to the guidelines established by the American Society of Echocardiography \cite{lang2015recommendations}. This annotation process was performed using Slicer 3D software \cite{fedorov20123d}, utilizing its manual segmentation tools for the delineation of anatomical structures.

\subsubsection{Diagnosis Step - Task 3 \& 4: AS Severity Detection and AR Severity Diagnosis}
The TMED-2 dataset \cite{wessler2023automated} comprises 599 fully labeled TTE studies from 577 patients as described in Task 1. The dataset includes B-mode imaging with five diagnosis labels: no AS, mild, mild to moderate, moderate, and severe AS. We follow the severity categorization of AS described in \cite{wessler2023automated} to evaluate three screening tasks: no AS, early AS (mild or mild to moderate) and significant AS (moderate or severe). We evaluated the model using a dataset divided into training, validation, and test sets consisting of 10,066, 3,602, and 3,602 samples, respectively.

The color Doppler imaging dataset for AR severity diagnosis \cite{kim2024assessment} comprises 183 cases from 250 patients at MGH. This dataset consists of 42 healthy cases and 141 diseased cases. Among these, 132 (72.1\%) were classified as non-severe cases, while the remaining 51 (27.9\%) were categorized as severe cases. Each study includes not only B-mode imaging but also multiple color Doppler imaging scans from different scan views: A3C, A5C, PSAX, and PLAX. Two cardiologists assessed and assigned severity grades to all patients using a scale ranging from 0.0 to 4.0 in 0.5 increments, corresponding to severity levels. A threshold of 2.5 was used to distinguish between severe and non-severe groups. 

In the data pre-processing stage, we resized all frames in the Echo dataset to 224 × 224 pixels and used 32 time frames to ensure consistency in the input size for our model. For model evaluation, we employ widely used metrics for segmentation evaluation, including the mean Dice coefficient (mDice), mean Intersection over Union (mIoU), Hausdorff Distance (HD), and Average Symmetric Surface Distance (ASSD). The standard deviations of these metrics were also reported. For view classification, AS detection, and AR severity diagnosis tasks, we use receiver operating curve analysis and report the area under the receiver operating characteristic curve (AUROC).

\subsection{Implementation details}
In our experiments, we implemented all ViT-based methods using ViT-L as our default setting. For patch embedding, we used $16 \times 16 \times 4$ patches, physical in-plane dimension of 2 cm/pixel through grid search. We selected a masking ratio of 75$\%$ for pretraining. For the downstream tasks, all images were resized to 224 $\times$ 224. Specifically, we employed data augmentation techniques for segmentation tasks, including random rotations, flips, translations, crops, and scaling. Specifically, rotations are applied within a range of \([-15^\circ, 15^\circ]\), while translations along the x and y dimensions are limited to \(\pm10\%\) of the image dimensions, and the scaling factors vary between 0.9 and 1.4 to include zoomed views which represent magnified scans of specific regions in the echocardiographic image. Random erasing is applied with a patch size of 16 \(\times 16\), which excludes certain parts of the object in the original image. For other downstream tasks, only horizontal and vertical flips are applied to maintain anatomical consistency, avoiding morphological changes that could affect diagnosis.



\begin{table}[t]
\centering
\caption{Quantitative results of various methods fine-tuned for the downstream view identification task. The accuracy of different methods on the TMED-2 and INTERNAL datasets is presented. The best result is highlighted in \textbf{bold}.}
\renewcommand{\arraystretch}{1.3}
\begin{tabular}{lccc}
\toprule[1.1pt]
    \multicolumn{1}{c}{\multirow{2}{*}{\textbf{Methods}}} & 
    \multicolumn{1}{c}{\multirow{2}{*}{\shortstack{\textbf{Pretrained}\\\textbf{Dataset}}}} &
    \multicolumn{2}{c}{\textbf{Accuracy (\%)}} \\
    \cmidrule(lr){3-4}
    & & \textbf{TMED-2} & \textbf{INTERNAL} \\
\hline
\hline
    ResNet50 \cite{he2016deep} & ImageNet & 83.7 & 78.4 \\
    ViT-B \cite{alexey2020image}  & ImageNet & 87.0  & 80.4 \\
    SAM \cite{kirillov2023segment}    & SA-1B & 92.3  & 86.3 \\
    LVM-MED \cite{mh2024lvm}          & 8 Medical imaging & 93.8 & 86.5 \\
    USFM \cite{jiao2024usfm}          & US & \textbf{94.3} & 86.2 \\
    VideoMAE \cite{tong2022videomae}  & Kinetics-400 \cite{kay2017kinetics} & 93.0  & 88.3 \\
\midrule
    EchoFM                           & - & 83.8 & 77.8 \\
    EchoFM                           & Countix \cite{wandt2019repnet} & 93.1 & 88.5 \\
    EchoFM (Ours)                          & Echo & 94.2 & \textbf{89.1} \\
\bottomrule[1.1pt]
\label{tab:view_identification}
\end{tabular}
\end{table}

\begin{table*}[h]
\caption{STATE-OF-THE-ART COMPARISON ON THE CAMUS DATASET. Quantitative results of methods fine-tuned for the chamber segmentation task on the CAMUS dataset. }
\renewcommand{\arraystretch}{1.2} 
\centering
\begin{tabularx}{\textwidth}{lXXXX XXXX XXXX |XXXX}
    \bottomrule[1.1pt]
  \multirow{3}{*}{\textbf{Methods}} & \multicolumn{4}{c}{\textbf{LV$_{endo}$}} 
    & \multicolumn{4}{c}{\textbf{LV$_{epi}$}} 
    & \multicolumn{4}{c}{\textbf{LA}} 
    & \multicolumn{4}{c}{\textbf{Average}} \\
    \cmidrule(lr){2-5} \cmidrule(lr){6-9} \cmidrule(lr){10-13} \cmidrule(lr){14-17}
    & Dice & IoU & HD95 & ASSD 
    & Dice & IoU & HD95 & ASSD
    & Dice& IoU & HD95 & ASSD
    & Dice & IoU & HD95 & ASSD
    \\
    \hline
    \hline
    LUNet \cite{leclerc2020lu} & 88.64 & 80.59 & 6.31 & 3.23 
    & 81.10 & 71.86 & 6.78 & 3.65 
    & 78.12 & 65.53 & 13.92 & 6.10 
    & 82.62 & 72.66 & 9.00 & 4.33 \\
    EchoNet \cite{ouyang2020video} & 89.83 & 80.31 & 6.08 & 2.98
    & 82.64 & 73.10 & 6.64 & 3.52
    & 79.51 & 67.60 & 11.02 & 5.64 
    & 83.99 & 73.67 & 7.91 & 4.05 \\
    SegFormer \cite{xie2021segformer} & 91.46 & 87.20 & 4.32 & 2.08
    & 88.18 & 82.10 & 4.85 & 3.15
    & 88.13 & \underline{85.90} & \underline{6.72} & 3.40 
    & 89.26 & 85.07 & 5.30 & 2.88 \\
    nnUnet ResEnc \cite{isensee2024nnu} & \underline{91.70} & \underline{89.01} & \underline{3.76} & 1.85 
    & \underline{88.24} & \underline{82.70} & 4.61 & 2.18 
    & 88.10 & 85.67 & 7.55 & 3.41 
    & \underline{89.35} & \underline{85.79} & 5.31 & 2.48 \\
    \midrule
    CLIP \cite{radford2021learning} & 87.45 & 78.82 & 7.12 & 3.33 
    & 81.65 & 73.10 & 5.84 & 3.23 
    & 79.82 & 68.87 & 11.04 & 4.96 
    & 82.97 & 73.60 & 8.00 & 3.84 \\
    VideoMAE \cite{tong2022videomae} & 91.80 & 85.16 & 5.51 & 2.56 
    & 86.31 & 77.05 & \underline{3.11} & 2.35 
    & 88.33 & 83.90 & \textbf{6.03} & 3.21 
    & 88.81 & 82.04 & 4.88 & 2.71 \\
    SimSiam \cite{chen2021exploring} & 90.18 & 88.08 & 5.90 & 3.44 
    & 85.88 & 78.05 & 4.65 & 2.02 
    & 87.82 & 82.90 & 7.82 & 4.05 
    & 87.96 & 83.01 & 6.12 & 3.17 \\
    DINO \cite{caron2021emerging} & 90.24 & 88.50 & 5.32 & 3.03 
    & 86.70 & 79.40 & 4.20 & \underline{1.38} 
    & 87.98 & 84.74 & 7.30 & 3.71 
    & 88.31 & 84.21 & 5.61 & 2.71 \\
    \midrule
    SAM \cite{kirillov2023segment} & 90.03 & 84.22 & 5.02 & 2.20 
    & 85.40 & 75.75 & 5.52 & 2.10 
    & 83.23 & 82.64 & 7.30 & 3.48 
    & 86.22 & 80.87 & 5.95 & 2.59 \\
    SonoSAM \cite{ravishankar2023sonosam} & 91.02 & 85.09 & 4.10 & 1.89 
    & 86.50 & 80.77 & 4.68 & 2.01 
    & 86.81 & 84.46 & 7.28 & 3.30 
    & 88.11 & 83.44 & 5.35 & 2.40 \\
    LVM-Med \cite{mh2024lvm} & 91.58 & 84.68 & 4.14 & \underline{1.77}
    & 85.78 & 76.20 & 4.72 & 1.77
    & 83.50 & 83.38 & 7.10 & 3.22 
    & 86.95 & 81.42 & 5.32 & 2.25 \\
    USFM \cite{jiao2024usfm} & 90.86 & 83.61 & 4.84 & 1.91 
    & 84.61 & 75.55 & 4.50 & 1.72 
    & 84.60 & 84.75 & 8.89 & 3.58 
    & 86.69 & 81.30 & 6.08 & 2.40 \\
    EchoCLIP \cite{christensen2024vision} & 91.30 & 88.94 & 4.06 & 2.24  
    & 87.98 & 82.10 & 3.89 & 1.92 
    & \underline{88.43} & 85.80 & 7.02 & \underline{3.20} 
    & 89.24 & 85.61 & 4.99 & 2.45 \\
    \midrule
    EchoFM (RandomInit) & 87.15 & 82.51 & 6.05 & 3.01 
    & 82.41 & 75.33 & 4.52 & 3.21 
    & 82.30 & 73.88 & 8.96 & 4.88 
    & 83.95 & 77.24 & 6.51 & 3.70 \\
    EchoFM (Countix) & 91.24 & 85.98 & 4.51 & 2.19 
    & 87.88 & 82.40 & 4.42 & 1.68 
    & 86.92 & 83.55 & 7.45 & 3.40 
    & 88.68 & 83.98 & \underline{4.79} & \underline{2.42} \\
    EchoFM (Ours) & \textbf{93.61} & \textbf{90.30} & \textbf{3.26} & \textbf{1.56}
    & \textbf{89.08} & \textbf{83.60} & \textbf{3.05} & \textbf{1.10}
    & \textbf{88.60} & \textbf{86.60} & \underline{6.05} & \textbf{2.77}
    & \textbf{90.43} & \textbf{86.83} & \textbf{4.12} & \textbf{1.81} \\
    \bottomrule[1.1pt]
\end{tabularx}
\label{table1}
\end{table*}
\begin{table*}[h]
\caption{
STATE-OF-THE-ART COMPARISON ON THE MULTI-CENTER DATASET. Quantitative results of methods fine-tuned for the chamber segmentation task on the Multi-center dataset.}
\renewcommand{\arraystretch}{1.2} 
\centering
\begin{tabularx}{\textwidth}{lXXXX XXXX XXXX |XXXX}
		\toprule
        \multirow{3}{*}{\textbf{Methods}} & \multicolumn{4}{c}{\textbf{LV$_{endo}$}} 
        & \multicolumn{4}{c}{\textbf{LV$_{epi}$}} 
        & \multicolumn{4}{c|}{\textbf{LA}} 
        & \multicolumn{4}{c}{\textbf{Average}} \\
        \cmidrule(lr){2-5} \cmidrule(lr){6-9} \cmidrule(lr){10-13} \cmidrule(lr){14-17}
        & Dice & IoU & HD95 & ASSD 
        & Dice & IoU & HD95 & ASSD
        & Dice& IoU & HD95 & ASSD
        & Dice & IoU & HD95 & ASSD 
        \\ 
        \hline
        \hline
        LUNet \cite{leclerc2020lu} 
        & 85.94 & 84.92 & 7.08 & 5.70 
        & 75.01 & 73.63 & 11.92 & 9.40 
        & 82.92 & 79.15 & 11.91 & 9.98 
        & 81.29 & 79.23 & 10.30 & 8.36 \\

        EchoNet \cite{ouyang2020video} 
        & 86.40 & 85.04 & 7.43 & 5.49 
        & 78.42 & 75.91 & 9.89 & 8.22 
        & 83.78 & 80.84 & 10.84 & 9.40 
        & 82.87 & 80.60 & 9.39 & 7.70 \\

        SegFormer \cite{xie2021segformer} 
        & 89.20 & 85.78 & 5.84 & 3.66 
        & 84.06 & 77.02 & 6.02 & \underline{4.08} 
        & 85.91 & 82.69 & 9.25 & 6.09 
        & 86.39 & 81.83 & 7.04 & 4.61 \\

        nnUnet ResEnc \cite{isensee2024nnu} 
        & 89.83 & 86.31 & 6.08 & \underline{2.98}
        & 82.64 & 75.10 & 6.64 & 4.16 
        & 85.51 & 83.60 & 9.02 & 5.64 
        & 85.99 & 81.67 & 7.25 & 4.26 \\

        \midrule
        
        CLIP \cite{radford2021learning} 
        & 83.19 & 81.28 & 10.21 & 7.19 
        & 79.19 & 75.94 & 9.04 & 8.13 
        & 83.14 & 80.90 & 14.87 & 10.94 
        & 81.84 & 79.37 & 11.37 & 8.75 \\

        VideoMAE \cite{tong2022videomae} 
        & 89.15 & 86.42 & 6.01 & 3.55 
        & 87.09 & 83.95 & \underline{5.29} & 4.40 
        & 85.59 & 84.22 & 10.32 & 4.66 
        & 87.28 & 84.86 & 7.21 & 4.20 \\

        SimSiam \cite{chen2021exploring} 
        & 87.80 & 85.90 & 7.50 & 4.51 
        & 86.88 & 84.05 & 6.05 & 5.01 
        & 84.20 & 83.33 & 11.22 & 5.23 
        & 86.29 & 84.43 & 8.26 & 4.92 \\

        DINO \cite{caron2021emerging} 
        & 89.09 & 86.34 & 6.30 & 4.03 
        & 86.98 & 83.80 & 5.74 & 4.80 
        & 84.90 & 84.20 & 11.32 & 5.20 
        & 86.99 & 84.78 & 7.79 & 4.68 \\
        
        \midrule
        
        SAM \cite{kirillov2023segment} 
        & 87.03 & 85.58 & 8.10 & 5.45 
        & 86.04 & 84.42 & 6.30 & 6.81 
        & 86.20 & 83.32 & 9.20 & 6.58 
        & 86.42 & 84.44 & 7.87 & 6.28 \\

        SonoSAM \cite{ravishankar2023sonosam} 
        & \underline{90.03} & \underline{87.22} & \underline{5.65} & 3.20
        & \underline{87.40} & \underline{83.75} & 5.52 & 4.16
        & 86.23 & \underline{84.64} & \underline{8.30} & \underline{4.48} 
        & \underline{87.89} & \underline{85.20} & \underline{6.49} & \underline{3.95} \\

        LVM-Med \cite{mh2024lvm} 
        & 88.01 & 86.03 & 7.98 & 4.88 
        & 86.12 & 85.19 & 7.12 & 6.20 
        & \underline{86.42} & 84.30 & 10.49 & 6.50 
        & 86.85 & 85.17 & 8.53 & 5.86 \\

        USFM \cite{jiao2024usfm} 
        & 86.80 & 85.24 & 7.33 & 3.74 
        & 85.88 & 81.32 & 6.34 & 5.29 
        & 85.10 & 83.04 & 11.23 & 8.02 
        & 85.93 & 83.20 & 8.30 & 5.68 \\

        EchoCLIP \cite{christensen2024vision} 
        & 88.70 & 86.36 & 5.82 & 3.52 
        & 87.03 & 83.38 & 5.60 & 4.30 
        & 86.08 & 83.24 & 10.01 & 5.24 
        & 87.27 & 84.33 & 7.14 & 4.35 \\
        \midrule
        EchoFM (RandomInit) 
        & 85.24 & 82.60 & 7.82 & 6.98 
        & 81.90 & 77.39 & 7.02 & 6.88 
        & 85.21 & 81.32 & 8.96 & 7.02 
        & 84.12 & 80.44 & 7.93 & 6.96 \\

        EchoFM (Countix) 
        & 89.82 & 86.98 & 6.20 & 3.55 
        & 86.05 & 81.96 & 6.88 & 5.81 
        & 85.78 & 82.87 & 9.54 & 5.60 
        & 87.22 & 83.94 & 7.54 & 4.99 \\

        EchoFM (Ours) 
        & \textbf{91.24} & \textbf{87.30} & \textbf{5.20} & \textbf{2.84} 
        & \textbf{88.53} & \textbf{85.48} & \textbf{4.56} & \textbf{3.90} 
        & \textbf{87.30} & \textbf{85.09} & \textbf{8.21}& \textbf{4.05} 
        & \textbf{89.02} & \textbf{85.96} & \textbf{5.99} & \textbf{3.60} \\

		\bottomrule[1.1pt]
\end{tabularx}
\label{table1-2}
\end{table*}

\subsection{Comparison with SOTA Methods Across Tasks}
\subsubsection{Acquisition Step - Task 1: Scan View Identification}
We compared our proposed method against baseline models and a variety of state-of-the-art methods.  The baseline models included ResNet50 \cite{he2016deep} and ViT-B \cite{alexey2020image}, both pretrained on the ImageNet-1K dataset. In addition, we evaluated the performance of VideoMAE \cite{tong2022videomae} and three foundation models including SAM \cite{kirillov2023segment}, LVM-Med \cite{mh2024lvm}, and USFM \cite{jiao2024usfm}. We utilized pretrained ViT blocks as the image encoder, paired with a lightweight decoder specifically optimized for classification tasks. Specifically, the encoder output is fused along the first dimension using average pooling, producing a size vector of features [1, D]. A single layer MLP is then used as the classification head, which maps the feature vector to its categorical representation of size [1, \textit{class}].

Table \ref{tab:view_identification} presents a performance comparison of various methods on the view identification task, evaluated using the TMED-2 dataset and an internal video dataset. We initialized the model with a pretrained image encoder and fine-tuned it on the public dataset. Our results showed that ResNet50 achieved lower accuracies of 83.7\% and 78.4\% on the public and internal datasets, respectively. In contrast, the ViT-based model demonstrated slightly more stable performance, achieving 87.0\% and 80.4\% on the respective datasets. Although USFM achieved the highest accuracy of 94.3\% on the TMED-2 dataset, EchoFM outperformed all other methods on the internal dataset with an accuracy of 89.1\%. This performance gap underscores the importance of temporal modeling in capturing the dynamic nature of cardiac motion. Unlike EchoFM, USFM does not incorporate mechanisms to effectively model temporal dependencies, limiting its ability to comprehensively analyze echocardiographic sequences.

Additionally, we evaluated the performance of EchoFM’s image encoder to assess the impact of pre-training datasets on its effectiveness. We compared EchoFM with three baseline models. These included a randomly initialized model, a model pretrained on the natural video dataset Countix \cite{wandt2019repnet}, and a model pretrained on a dedicated echocardiography dataset. The evaluation shows that EchoFM, leveraging domain-specific pretraining on echocardiographic data, consistently surpassed both the randomly initialized and natural video-pretrained models on the TMED-2 and internal datasets. Notably, EchoFM achieved a 0.9\% and 0.6\% improvement in accuracy over the Countix-pretrained model in public and internal datasets, respectively, underscoring the advantages of domain-specific pretraining.

\begin{figure*}[h]
    \centering
    \includegraphics[width=1.0\linewidth]{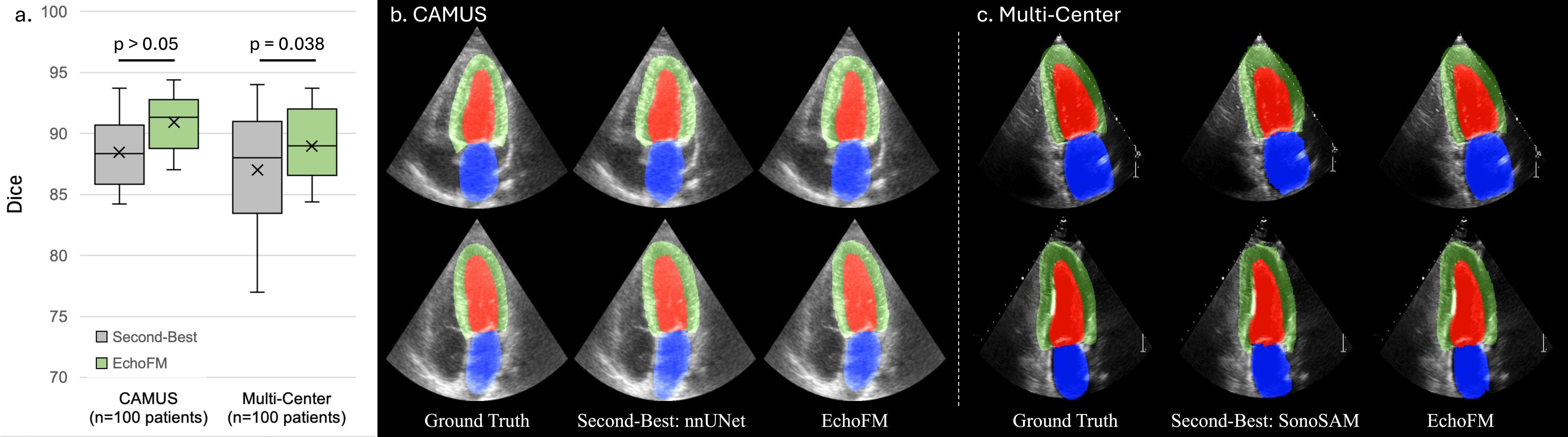}
    \caption{(a) Quantitative comparison of Echocardiography segmentation performance in the Dice metric. The Dice metric for each trial is presented with box-and-whiskers plot representing the range from minimum to maximum values. The p values indicate the statistically
significant superiority of the proposed model. All statistical tests were two-sided. Visual comparison of EchoFM and the Second-Best model in CAMUS (b) and Multi-center dataset (c).}
    \label{vis}
\end{figure*}

\subsubsection{Measurement and Analysis Step - Task 2: Chamber Segmentation}
For chamber segmentation tasks, we compared our method with several state-of-the-art medical image segmentation techniques. The baseline models included LUNet \cite{leclerc2020lu}, which integrates localization and segmentation methods sequentially, and EchoNet-Dynamic \cite{ouyang2020video}, which leverages 2D spatial and 1D convolutional neural networks. Furthermore, we evaluated CNN-based methods, including nnU-Net \cite{isensee2024nnu}, a self-configuring framework built upon the U-Net architecture, widely recognized for its strong performance in 3D medical image segmentation challenges. Additionally, we assessed SegFormer \cite{xie2021segformer}, which integrates a ViT encoder pre-trained on ImageNet with a specialized decoder tailored for segmentation tasks. We also evaluated several self-supervised learning (SSL) methods, including CLIP \cite{radford2021learning}, VideoMAE \cite{tong2022videomae}, SimSiam \cite{chen2021exploring}, and DINO \cite{caron2021emerging}. Due to the absence of paired text descriptions in our dataset, we used a pretrained CLIP vision encoder pretrained on natural images. This limitation makes pretraining CLIP on echocardiography data infeasible, whereas the other SSL methods are pretrained directly on echocardiography data. We included five foundation models, SAM \cite{kirillov2023segment}, SonoSAM \cite{ravishankar2023sonosam}, LVM-Med \cite{mh2024lvm}, USFM \cite{jiao2024usfm}, and EchoCLIP \cite{christensen2024vision}. The pretrained image encoders were paired with a decoder architecture inspired by TransUnet \cite{chen2021transunet}, incorporating a transformer-based two-dimensional approach built upon the U-Net framework. For video models, we adopted a decoder architecture based on UNETR \cite{hatamizadeh2022unetr}, specifically designed for three-dimensional transformer-based segmentation tasks.

Table \ref{table1} and Table \ref{table1-2} present comparative results for chamber segmentation tasks, focusing on LV${endo}$, LV${epi}$, and LA. Table \ref{table1} summarizes the results on the public dataset, and Table \ref{table1-2} highlights performance on the multi-center dataset. In the evaluation of the CAMUS dataset, we observed that EchoFM outperforms the second-best approach, nnU-Net, achieving an average dice score of 90.43 (95\% CI 87.21-93.65), showing an average of 1.08 dice improvement. Pretraining on large-scale, domain-specific datasets has consistently demonstrated the ability to enhance performance on unseen datasets, underscoring the superior generalization capabilities of such models to out-of-distribution data. To evaluate this, we conducted experiments using a multi-center dataset. EchoFM exhibited superior performance compared to the second-best method, SonoSAM, achieving an average Dice score of 89.02 (95\% CI: 85.35–92.69), with a notable improvement of 1.13 in the average Dice score. 

To further examine the effect of pretraining datasets on performance in unseen datasets, we evaluated EchoFM under three configurations: with random initialization, pretraining on the Countix dataset, and pretraining on an echocardiography-specific dataset. EchoFM pretrained on the echocardiography dataset achieved the highest average Dice score of 89.02\%, surpassing its randomly initialized counterpart and the Countix-pretrained model by 4.9\% and 1.8\%, respectively. These findings support the critical importance of leveraging domain-specific data for pretraining to optimize model performance on downstream tasks. In Fig. \ref{vis} (a), while the overall Dice score improved on the CAMUS dataset, the difference was not statistically significant. However, EchoFM demonstrated significant improvement on the unseen multi-center dataset, highlighting its superior generalization capability. Fig. \ref{vis} (b,c) provides a visual comparison between the second-best model and EchoFM, illustrating the qualitative enhancements achieved by EchoFM in segmentation tasks. To assess clinical applicability, we evaluated the left ventricular ejection fraction (LVEF) across models. EchoFM demonstrated superior performance, achieving an R² value of 0.86, significantly outperforming the second-best model, which achieved an R² value of 0.78, on the multi-center dataset.

\begin{table}[t]
\centering
\caption{
The performance of different methods fine-tuned on the TMED-2 for AS detection. AUROC is reported.}
\renewcommand{\arraystretch}{1.3}
\begin{tabularx}{\linewidth}{>{\centering\arraybackslash}m{0.21\linewidth} 
>{\centering\arraybackslash}m{0.13\linewidth} 
>{\centering\arraybackslash}m{0.15\linewidth} 
>{\centering\arraybackslash}m{0.15\linewidth} 
>{\centering\arraybackslash}m{0.15\linewidth}}
\toprule[1.1pt]
    \textbf{Methods} & \textbf{Pretrained Dataset} & \textbf{Absent vs. Present} & \textbf{Early vs. Significant} & \textbf{Non vs. Significant} \\
\midrule
\midrule
    ResNet50 \cite{he2016deep} & ImageNet & 0.66 & 0.67 & 0.66 \\
    ViT-B \cite{alexey2020image}  & ImageNet           & 0.68                        & 0.71                           & 0.70 \\
    AAD \cite{wessler2023automated}   & --                  & 0.74                      & 0.73                         & 0.76                        \\
    
    SAM \cite{kirillov2023segment}    & SA-1B              & 0.74                      & 0.76                         & 0.77                        \\
    LVM-MED \cite{mh2024lvm} & 8 Medical Imaging & 0.76                      & 0.77                         & 0.79                        \\
    USFM \cite{jiao2024usfm}          & US                 & 0.77                      & 0.80                         & 0.81                        \\
    EchoCLIP \cite{christensen2024vision} & Echo        & 0.79                      & 0.82                         & 0.82                        \\
\midrule
    EchoFM                            & --                  & 0.72            & 0.71               & 0.70            \\
    EchoFM                 & Countix             & 0.75            & 0.76               & 0.77              \\
    EchoFM                     & Echo                & \textbf{0.81}            & \textbf{0.84}               & \textbf{0.83}              \\
\bottomrule[1.1pt]
\end{tabularx}
\label{tab:AS_three}
\end{table}

\subsubsection{Diagnosis Step - Task 3 $\&$ 4: AS and AR Severity Diagnosis}
In the diagnosis step, we established two downstream tasks: evaluating the severity of aortic stenosis (AS) using B-mode imaging and aortic regurgitation (AR) using Color Doppler imaging. We compared our proposed method against baseline models and a variety of state-of-the-art methods. For the AS severity detection task, we evaluated our proposed method against baseline models and various state-of-the-art methods. The baseline models included ResNet50 \cite{he2016deep} and ViT-B \cite{alexey2020image}, both pretrained on the ImageNet-1K dataset. Additionally, we compared our method with the Automated AS Detection Method (AAD) based on \cite{wessler2023automated} and several foundation models, including SAM \cite{kirillov2023segment}, LVM-Med \cite{mh2024lvm}, USFM \cite{jiao2024usfm}, and EchoCLIP \cite{christensen2024vision}. Following the approach used in the view identification task, the encoder output is aggregated along the first dimension using average pooling, resulting in a feature vector of size [1, D]. This vector is then passed through a single-layer MLP, serving as the classification head, to map the features to their categorical representation of size [1, \textit{class}].

\begin{table}[t]
\centering
\caption{Quantitative results of various methods fine-tuned for the downstream AR severity diagnosis task, evaluated in terms of accuracy, precision, recall, and F1 score on the internal dataset. The best result is highlighted in \textbf{bold}.}

\renewcommand{\arraystretch}{1.3}
\begin{tabularx}{\linewidth}{>{\centering\arraybackslash}m{0.22\linewidth} 
>{\centering\arraybackslash}m{0.13\linewidth} 
>{\centering\arraybackslash}m{0.1\linewidth} 
>{\centering\arraybackslash}m{0.1\linewidth} 
>{\centering\arraybackslash}m{0.1\linewidth}
>{\centering\arraybackslash}m{0.08\linewidth}}
\toprule[1.1pt]
\textbf{Methods} & \textbf{Pretrained Dataset} & \textbf{Accuracy} & \textbf{Precision} & \textbf{Recall} & \textbf{F1} \\
\midrule   
\midrule
MVCN \cite{kim2024assessment} & Kinetics400 & 83.3 & 86.4 & 82.2 & 84.2 \\
VideoMAE \cite{tong2022videomae} & Echo & 84.6 & 87.0 & 86.1 & 86.5 \\
\midrule
EchoFM & -- & 70.8 & 72.3 & 69.8 & 71.0 \\
EchoFM & Countix & 85.1 & 86.9 & 86.4 & 86.6 \\
EchoFM (Ours) & Echo & \textbf{85.7} & \textbf{87.1} & \textbf{86.6} & \textbf{86.8} \\
\bottomrule[1.1pt]    
\end{tabularx}
\label{tab:AR_1}
\end{table}

Table \ref{tab:AS_three} presents the AUROC results for various methods, evaluated based on three specific AS screening tasks. The five levels of severity of AS were grouped into binary classifications for this analysis. Screening tasks include (i) distinguishing between absent and present AS (any severity), (ii) differentiating between early AS (mild or mild-to-moderate) and significant AS (moderate or severe), and (iii) distinguishing nonsignificant AS (none, mild, or mild-to-moderate) from significant AS (moderate or severe). We first compared EchoFM with baseline models, including ResNet50 and ViT-B, and observed significant performance improvements. EchoFM achieved an average AUROC improvement of 0.16 and 0.13 over ResNet50 and ViT-B, respectively. Furthermore, EchoFM outperformed the second-best method, EchoCLIP, with improvements of 0.2, 0.2, and 0.1 across all screening tasks, highlighting its robust capabilities. Consistent with previous tasks, EchoFM pretrained on a domain-specific dataset demonstrated superior performance compared to models with random initialization or pretrained on natural image datasets.

For the AR severity diagnosis task, which relies inherently on video data, our evaluation was focused exclusively on video-based models to ensure robust and meaningful comparisons. The Multiview Video Contrastive Network (MVCN) \cite{kim2024assessment} served as the baseline model for this task. MVCN employs a video encoder to extract high-level features from multiple echocardiographic views, integrating these representations through contrastive learning. The original MVCN utilized a video encoder pretrained on natural images. To investigate the impact of domain-specific pretraining, we replaced the video encoder with models pretrained on VideoMAE \cite{tong2022videomae} and EchoFM.

The diagnostic performance for AR severity across A3C, A5C, PSAX, and PLAX views is summarized in Table \ref{tab:AR_1}. VideoMAE achieved an accuracy of 84.6\%, surpassing the baseline MVCN’s accuracy of 83.3\%. Incorporating the EchoFM-pretrained encoder further enhanced the performance, achieving an accuracy of 85.7\%. These results underscore the effectiveness of EchoFM in extracting domain-specific features, offering superior representation learning tailored for echocardiographic tasks.

\subsection{Ablation Studies}
\label{sec:ablation}
In this section, we conducted comprehensive ablation experiments to investigate the performance of various factors that may impact downstream task performance, including (i) EchoFM without using periodic contrastive learning in Eq. \ref{eq:loss2}, which enforces periodic representation in latent space; (ii) label efficiency; (iii) parameter analysis that includes the spatial-temporal token size and masking ratio.

\begin{figure}
\centering
	\includegraphics[width=0.35\textwidth]{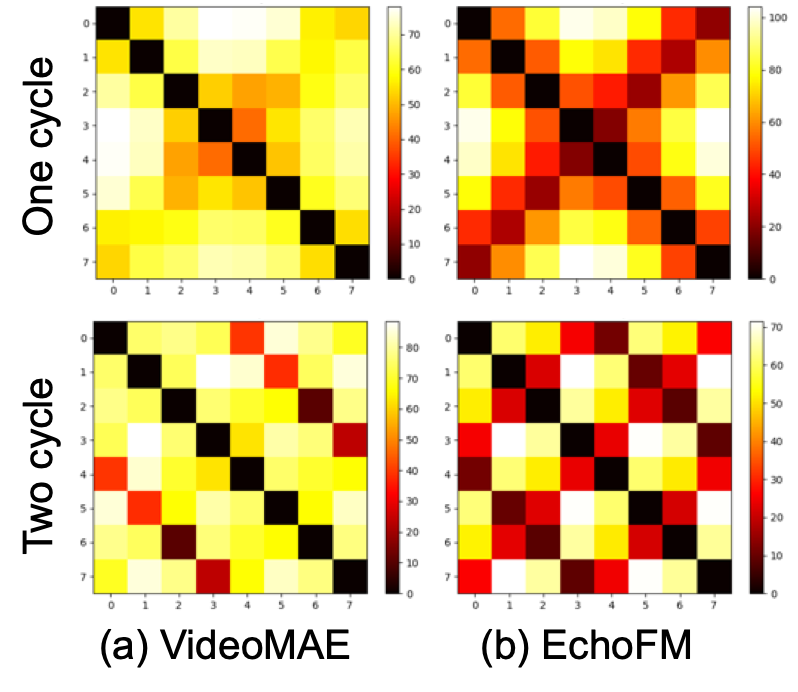}
	\caption{Comparison of temporal self-similarity matrices produced by two methods: (a) VideoMAE and (b) the proposed EchoFM, illustrating the temporal self-similarity matrix within one and two cardiac cycles. }
	\label{fig:periodic}
\end{figure}

\subsubsection{Effectiveness of periodic contrastive learning}
\begin{figure}[!t]
\centering
	\includegraphics[width=0.450\textwidth]{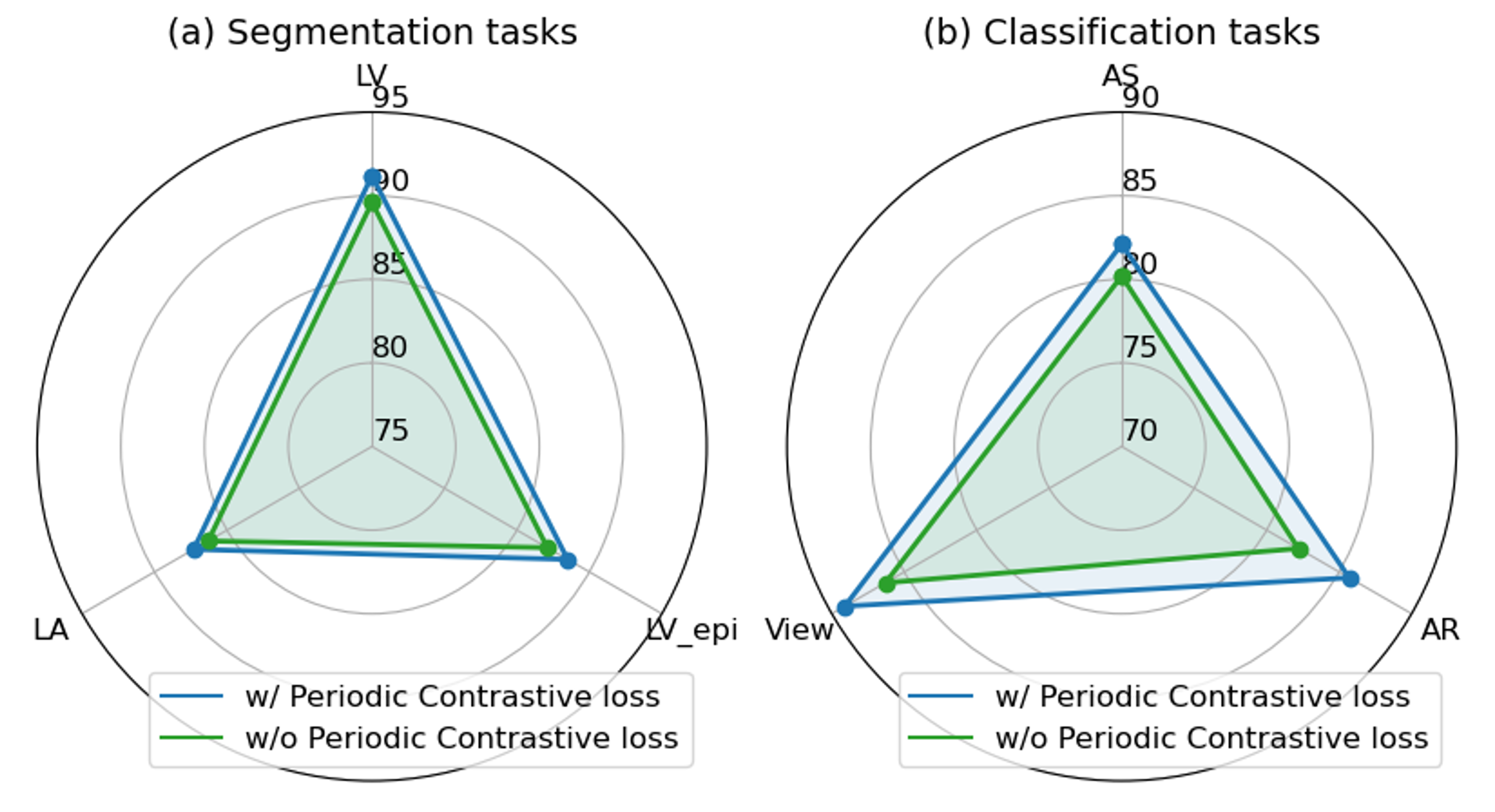}
	\caption{Ablation study evaluating the impact of periodic contrastive learning on model performance. (a) Dice scores for the chamber segmentation task. (b) Accuracy results for classification downstream tasks. Models trained with periodic contrastive loss are represented in blue, while models trained without it are shown in green.}
	\label{fig:radar}
\end{figure}

We investigated the impact of a key component of our method, periodic contrastive loss, by comparing temporal self-similarity matrices across multiple cardiac cycles, as illustrated in Fig. \ref{fig:periodic}. A clear periodic pattern emerged in the pretrained model with periodic contrastive loss, as observed by comparing the first and second columns. Figure \ref{fig:radar} demonstrates the performance gains achieved by EchoFM with periodic contrastive loss in the segmentation and classification tasks. For segmentation, PCL improved the dice scores for LV$_{endo}$ (91.2 vs. 89.6), LV$_{epi}$ (88.5 vs. 87.1), and LA (87.3 vs. 86.3). Similarly, in classification tasks, PCL enhanced accuracy for AS detection (82.1\% vs. 80.2\%), AR severity estimation (85.7\% vs. 82.2\%), and view identification (89.1\% vs. 86.3\%). These results underscore the effectiveness of PCL in improving task performance. We observed that models initialized with pretrained weights consistently outperformed those with random initialization, demonstrating the benefits of leveraging pretrained representations for enhanced performance. Specifically, we initialized the encoder with pretrained VideoMAE trained on echocardiography, which facilitates better optimization of periodic contrastive loss, as shown in Fig. \ref{fig:periodic} (a). 

\subsubsection{Label efficiency}

\begin{figure}[!t]
\centering
	\includegraphics[width=0.4\textwidth]{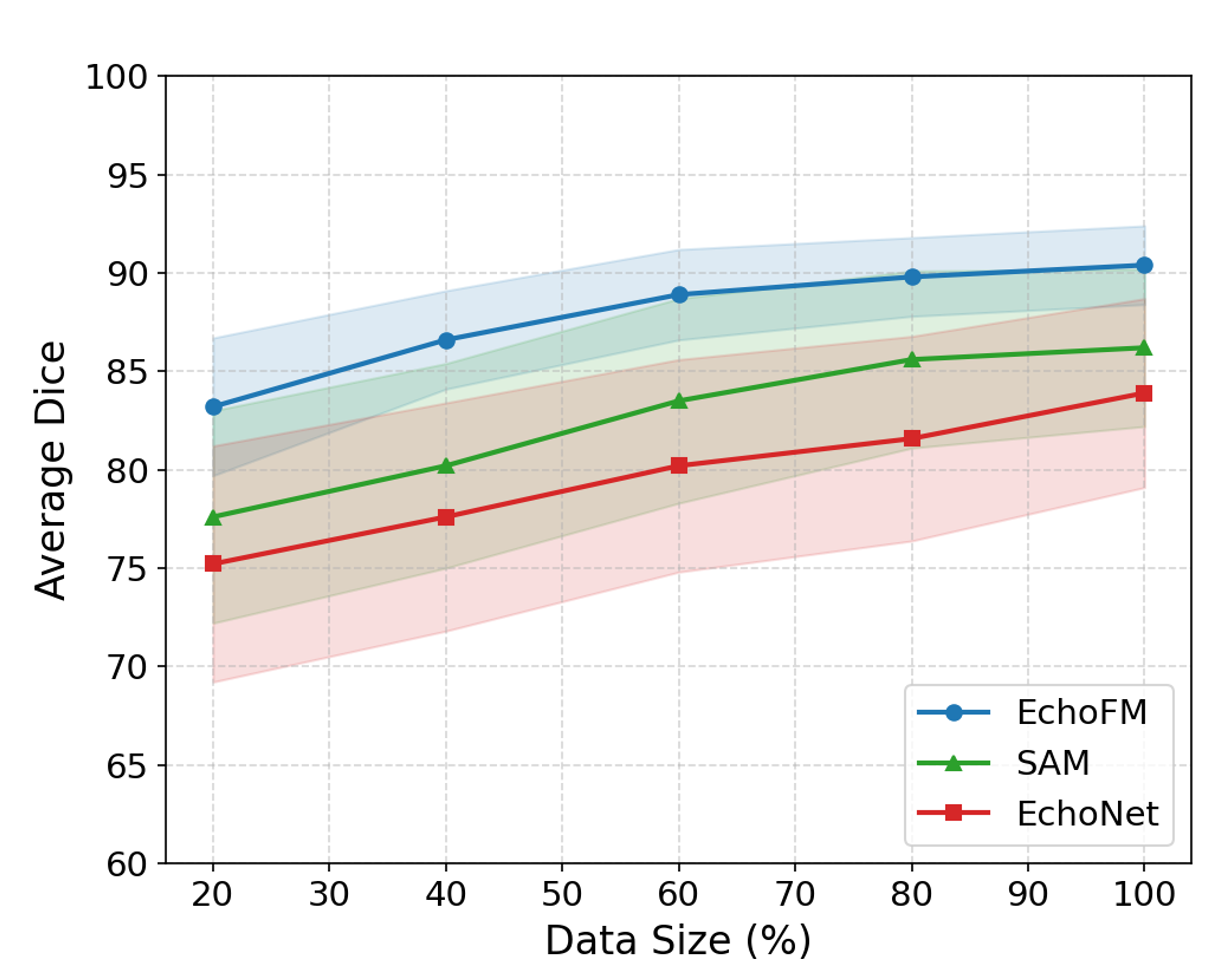}
	\caption{Comparison of various methods on the chamber segmentation task, focusing on label efficiency. EchoFM, when compared with EchoNet without pre-training and SAM pretrained on natural images, exhibits superior performance, particularly in scenarios with very limited labeled data. The results are presented with 95\% confidence intervals.}
	\label{fig:data_efficiency}
\end{figure}

One of the key advantages of a foundation model in the medical domain is its efficiency in labeling. We compare the label efficiency of the model on chamber segmentation tasks using varying amounts of labeled data for training. We include three models in our comparison: EchoNet without pre-training, SAM pretrained on natural images, and EchoFM pretrained on echocardiography data. In Fig. \ref{fig:data_efficiency}, the results show that EchoFM consistently outperforms other models at all levels of data use. Remarkably, even with extremely limited labeled data, EchoFM achieves an average Dice score of 83.2\%, which is nearly equivalent to EchoNet's 83.9\% score obtained using fully labeled data. Furthermore, our observations indicate that supervised methods without pre-training on large-scale datasets are heavily dependent on the availability of annotated data. These results highlight that EchoFM not only significantly outperforms other models, especially those without pre-training, but also demonstrates remarkable label efficiency.

\subsubsection{Parameter analysis}
We conducted ablation experiments to analyze the impact of spatio-temporal patch size on model performance. As shown in Table \ref{ab:patch_size}, models exhibited stable performance when trained with a high masking ratio up to 75\%, with minimal performance degradation observed in masking 90\%. Increasing the in-plane patch size from 8 to 16, which corresponds to 2 cm per patch, resulted in notable improvements in Dice scores. This highlights the benefits of larger patch sizes for capturing detailed spatial information. Additionally, we evaluated the effect of temporal size by comparing configurations with 2 and 4 frames. The results demonstrated that the optimal performance was achieved with 4 frames when combined with an in-plane patch size of 16.

\section{Discussion} 
\label{sec:discussion}
In this work, we proposed EchoFM, a general-purpose foundation model for echocardiography, trained on a large-scale dataset over 20 million echocardiographic image from 6500 patients. EchoFM leverages masked image modeling combined with periodic learning, utilizing a temporal self-similarity matrix with triplet loss to effectively capture spatio-temporal representations. To efficiently learn periodic representations, we introduced a spatio-temporal consistent masking strategy that enhances periodic contrastive learning. The periodic learning mechanism employed by EchoFM leverages the cyclic nature of cardiac motion, allowing the model to capture and utilize temporal dependencies effectively. 

Our experiments demonstrate EchoFM’s superior performance in various downstream tasks, such as view identification, chamber segmentation, and disease classification. EchoFM consistently outperforms existing methods, achieving higher accuracy and efficiency. For example, in view identification tasks, EchoFM demonstrates higher classification accuracy by effectively distinguishing between different echocardiographic views. In chamber segmentation, it achieves superior dice coefficients, indicating a more precise delineation of cardiac structures compared to existing models. In disease classification, such as the detection of aortic stenosis, EchoFM shows enhanced sensitivity and specificity, ensuring better diagnostic performance. Compared to the recently proposed EchoPrime model, a multi-view video-informed framework \cite{fan2021multiscale}, EchoFM demonstrated superior performance in estimating left ventricular ejection fraction (LVEF), achieving an R² value of 0.84 compared to EchoPrime’s 0.79.

Despite EchoFM's outstanding performance across various tasks, there is one limitation that could affect its ability to understand complex echocardiography videos. The performance of the model is highly dependent on the diversity of training data. Dataset biases, particularly those involving rare or atypical cardiac events, could potentially impact the robustness of EchoFM. For example, atrial fibrillation, characterized by irregular heart rhythms, presents challenges in capturing consistent periodic representations. This irregularity hinders the effectiveness of our periodic contrastive learning approach and remains an area that requires further exploration.

Future work will focus on expanding the training datasets to encompass a broader spectrum of cardiac conditions, including rare and atypical cases, to further enhance the diagnostic capabilities of the model. Furthermore, integrating multimodal data sources, such as combining echocardiographic imaging with patient clinical data, could provide a more holistic approach to the assessment of cardiac health. In addition, continued refinement of the model learning algorithms to better handle diverse and challenging imaging conditions will be critical to pushing the boundaries of what is achievable with artificial intelligence in medical imaging. Future research and development will further enhance the capabilities and applications of EchoFM, shedding light on its potential for a more effective and efficient use of echocardiography data in healthcare.

\begin{table}[t]
\centering
\caption{Ablation study of the proposed EchoFM foundation model in terms of token size and masking ratio. The best result is shown in \textbf{bold}.}
\renewcommand{\arraystretch}{1.3}
\begin{tabular}{c@{\hspace{0.05cm}}c}
    \begin{minipage}{0.48\linewidth}
    \centering
    \begin{tabular}{c|c}
        \bottomrule[1.1pt]
        Patch Size & mDice [\%] \\
        \hline   
        \hline   
        8 $\times$ 8 $\times$ 2   & 89.20 \\
        8 $\times$ 8 $\times$ 4   & 89.03 \\
        16 $\times$ 16 $\times$ 2 & 89.65 \\
        16 $\times$ 16 $\times$ 4 & \textbf{90.43} \\
        \bottomrule[1.1pt]
    \end{tabular}
    \end{minipage}
    &
    \begin{minipage}{0.48\linewidth}
    \centering
    \begin{tabular}{l|c}
        \bottomrule[1.1pt]
        Ratio & mDice [\%] \\
        \hline   
        \hline   
        25 \% & 89.02 \\
        50 \% & 89.46  \\
        75 \% & \textbf{90.43} \\
        90 \% & 90.20 \\
        \bottomrule[1.1pt]  
    \end{tabular}
    \end{minipage}
    \label{ab:patch_size}
\end{tabular}
\end{table}

\section{Conclusion}
\label{sec:Conclusion}
EchoFM represents a significant advancement in echocardiography analysis, demonstrating strong generalizability and superior performance in various downstream tasks. EchoFM, trained on a comprehensive echocardiography dataset comprising over 290,000 videos from 6,500 patients with diverse clinical conditions, effectively addresses the challenges of echocardiography and demonstrates outstanding performance across various downstream tasks. The model employs an innovative approach that incorporates the reconstruction of missing tokens, periodic learning with a temporal self-similarity matrix, and spatio-temporal consistent masking, enabling robust and accurate analysis. Extensive experiments demonstrate that EchoFM surpasses existing methods in view identification, chamber segmentation, and disease classification both in public and inhouse dataset aligning with routine echocardiographic workflows. This highlights its potential to enhance clinical practice and contribute to better patient outcomes.

\bibliographystyle{IEEEtran} 
\bibliography{IEEEtran}


\end{document}